\documentclass[10pt,twocolumn,letterpaper]{article}
\usepackage{3dv}
\usepackage{times}
\usepackage{epsfig}
\usepackage{graphicx}
\usepackage{amsmath}
\usepackage{amssymb}
\usepackage{cite}
\usepackage{subfig}
\usepackage{amsmath}
\usepackage{algorithmic,algorithm}

\threedvfinalcopy 

\def\threedvPaperID{95} 

\ifthreedvfinal\fi
\begin{document}

\title{Multi-View Inpainting for RGB-D Sequence}

\author{
	Feiran Li, Gustavo Alfonso Garcia Ricardez, Jun Takamatsu and Tsukasa Ogasawara\\
	Nara Institute of Science and Technology\\
    8916-5 Takayama-cho, Ikoma, Nara, Japan\\
	{\tt\small \{li.feiran.kw8,	garcia-g, j-taka, ogasawar\}@is.naist.jp}
}

\maketitle

\begin{abstract}
In this work we propose a novel approach to remove undesired objects from RGB-D sequences captured with freely moving cameras, which enables static 3D reconstruction. Our method jointly uses existing information from multiple frames as well as generates new one via inpainting techniques. We use balanced rules to select source frames; local homography based image warping method for alignment and Markov random field (MRF) based approach for combining existing information. For the left holes, we employ exemplar based multi-view inpainting method to deal with the color image and coherently use it as guidance to complete the depth correspondence. Experiments show that our approach is qualified for removing the undesired objects and inpainting the holes. 
\end{abstract}


\section{Introduction}
3D reconstruction tools~\cite{newcombe2011kinectfusion,kerl2013dense} such as structure from motion (SfM) and visual simultaneous localization and mapping (v-SLAM) serve for many purposes from path planning to scene understanding. Many of such approaches have provided RGB-D versions for the easily achievable distance information. However, with these classical methods undesired objects would be inevitably introduced to the constructed map if the input sequences are captured in dynamic environments. Some methods~\cite{caccamo2018joint,ataer2016object} present solutions to deal with the rigidly moving objects by clustering features into either static or dynamic classes. It was not until the recent blossom of image semantic segmentation that demonstrates new insight on how to remove the undesired objects in 2D image level with more flexibility.

However, simply removing the unwanted objects would leave blanks on the images and hence we focus on filling in these holes with inpainting techniques. Within the off-the-shelf approaches, the single image inpainting methods~\cite{barnes2009patchmatch,kawai2009image} are useless to handle large holes since the source region on a single image cannot provide enough information. As for the video completion algorithms~\cite{ebdelli2015video,klose2015sampling}, they are specifically designed to deal with color videos and hence not suitable for the RGB-D sequences in our case. We consider our method as one of the multi-view inpainting approaches, which are much more flexible than video completion for being able to handle both color and RGB-D images as well as to make use of information gotten from various kinds of sources, such as images captured with stereo cameras or the discrete ones.
 
In this paper we propose a unified framework to remove robust classes of clutters from RGB-D sequences. With a slight abuse of notation we inherently define a pair of color and depth images as a ``frame", which is the minimum unit in our algorithm. For a certain frame with masks to fill in, our algorithm searches useful information from the other counterparts. We consider our system as semi-automatic since one can conveniently remove the annoyances via defining their semantic classes.

We summarize the contributions of this paper as follows. First, we introduce a source selection strategy which carefully balances the internal trade-offs among distinct frames to attain expected results. Second, we use the local homography based warping method to achieve more accurate alignments than previous work~\cite{whyte2009get,ebdelli2015video}; as well as to prevent information loss during 2D-3D projection in the SfM based methods~\cite{thonat2016multi,philip2018plane}. Our third contribution is a series of inpainting methods to make use of information from multiple source frames, in which we propose an MRF based approach for combining the candidates; extend the searching region of exemplar based color image inpainting methods to multiple views and coherently inpaint the depth.

The remainder of this paper is organized as follows. Section~\ref{relatedWork} makes a brief introduction of related inpainting work. Section~\ref{overview} gives an overview of our entire algorithm. Section~\ref{sourceFrameSelection} introduces the strategy to select source frames. Section~\ref{registerAndCombine} and~\ref{holeFilling} describe the specifics to achieve color and depth inpainting. Section~\ref{resultAndCompare} shows the tests and comparison results. Finally, the conclusions are given in section~\ref{conclusionAndDiscussion}.


\section{Related Work}\label{relatedWork}
Image inpainting contains broad research fields and hereby we briefly review those closely related to our work. Our approach is a combination of multi-view image inpainting and depth Inpainting.

\subsection{Single Image Inpainting}
Exemplar based methods are popular in single image inpainting for their texture-protection ability. The beginning of them can be traced to the work of Criminisi \textit{et al.}~\cite{criminisi2004region}, in which the mask is completed via searching for similar patches from the rest region and inherently copying them. The PatchMatch algorithm proposed by Barnes \textit{et al.}~\cite{barnes2009patchmatch} uses random search for quickly finding approximate nearest neighbor matches between patches, which is widely employed as the basis in the follow-up work for its several orders higher time efficiency. Kawai \textit{et al.}~\cite{kawai2009image} extend the energy function by taking into account of brightness changes and spatial locality of texture to deal with unnatural matches. Lee \textit{et al.}~\cite{lee2016laplacian} propose to take Laplacian pyramid as an error term in patch synthesis in order to protect edges. In summary, the single image inpainting approaches leverage information from the image itself. In contrast, we use the other frames as additional sources.   

\subsection{Video Completion}
Video completion aims at dealing with color image sequences. Some work requires manual interaction: Klose \textit{et al.}~\cite{klose2015sampling} propose to inpaint a given video by using SfM and manually drawing 3D masks. Other methods either completely copy information from other frames or generate new textures by searching from them: Granados \textit{et al.}~\cite{granados2012background} enable free movement of camera via using multiple homographies to estimate the geometric registration between frames; whose applications are limited for being required to satisfy the assumption that the missing pixels on the target frame can be completely achieved from the others. On the other hand, Newson \textit{et al.}~\cite{newson2014video} propose an exemplar based method to search for similar patches on a group of aligned source frames, which is pretty time consuming for minimizing a global energy function. Similarly, Ebdelli \textit{et al.}~\cite{ebdelli2015video} shrink the searching range by only considering a small number of aligned neighboring frames of the target one. Differently with these methods that handle color videos, our approach is designed for RGB-D sequences; Also we take advantages from both direct copying and multi-view searching and hence more suitable for highly textured scenes.

\subsection{Multi-View based Inpainting}
Multi-view inpainting techniques leverage information from multiple source frames. Hays and Efros~\cite{hays2007scene} gather photos from Internet as a huge database to help with image completion. Similarly, Whyte \textit{et al.}~\cite{whyte2009get} cover an undesired region on the query image with Internet photographs of the same scene, in which multi-homography and photometric registration are used to achieve geometric registration between the query image and the source ones. Also a Markov random field optimization~\cite{agarwala2004interactive} is employed for selecting the optimal proposals. This kind of methods are pretty unstable since the masked objects are easy to be reintroduced as a result of lacking necessary means to filer the source information.

Recent research begins to show interests on using the geometric connections among different views. Baek \textit{et al.}~\cite{baek2016multiview} present a multi-view based method to complete the user-defined region by jointly inpaint the color and depth image, which takes advantages from SfM to achieve geometric registration among different views. Similarly, Thonat \textit{et al.}~\cite{thonat2016multi} enable free-viewpoint image based rendering with reprojected information from neighboring views. Also a refined method is proposed in the following work~\cite{philip2018plane} that performs inpainting on intermediate, local planes in order to preserve perspective as well as to ensure multi-view coherence. 
In contrast, we use local homography for achieving pixel-wise correspondences, with which the information loss caused by SfM could be effectively avoided. Also they assume that the input images are of high quality, while in contrast, our approach aims at dealing with more common scenarios, such as those taken with moving cameras and hence glutted with blurriness.   

\subsection{Depth Inpainting}
Depth inpainting is similar to the propagation methods designed for the color images to a certain degree. Result quality may however be limited if the algorithms designed for color images are simply transplanted to the depth counterparts. Therefore popular solutions use color images as guidance to complete the holes on the depth ones. Miao \textit{et al.}~\cite{miao2012texture} introduce a texture assisted inpainting technique via dividing the target area into smooth and edge classes and distribute different partial differential equations (PDE) to each class. Atapour-Abarghouei \textit{et al.}~\cite{atapour2017depthcomp} perform semantic segmentation on the color images to get the object edges and the depth value is coherently propagated within every object. Such work targets on assigning value to each unknown pixel. In this work, however, we take the unknown as one of the existing values and only inpaint the mask left by the removed undesired objects.   


\begin{figure*}[t]
\centering
  \includegraphics[width=\textwidth]{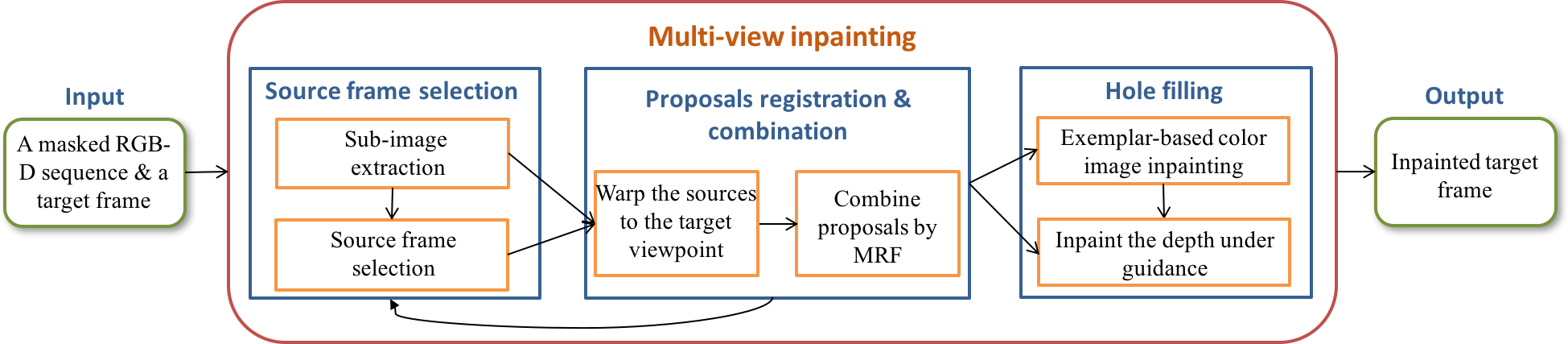}
  \caption{Overview of our method. For every input target frame, the first step is to select satisfactory source frames for inpainting. Then for every sub-images (specifically described in section~\ref{localSimilarity}), information from the corresponding sources is comprehensively combined. Finally the left holes on the color and depth images are inpainted separately.}
  \label{pipeline}
\end{figure*}

\section{Overview}\label{overview}
The proposed pipeline can be found in Fig.~\ref{pipeline}. A masked RGB-D sequence \(F\) and one target frame \(F_{t}\) are taken as input. The RGB-D sequence serves as source frames from where inpainting information can be gotten and the masks stand for the objects that we would like remove, which are semi-automatically generated in our work with the help of deep learning based semantic segmentation techniques. Our goal is to fill in the masks on \(F_{t}\) with realistic content by using multi-view information. In order to achieve it, we carefully select a set of source frames from \(F\) for \(F_{t}\). Also we take benefits from local homography based image warping method~\cite{zaragoza2013projective} to warp each source \(F_{i}\) into the same image coordinate with \(F_{t}\). Considering that the warpings are not equally accurate, we use a MRF approach similar to those proposed in~\cite{agarwala2004interactive,whyte2009get} to reduce bias. After this, we separately use exemplar based multi-view inpainting algorithm to cope with the color image \(I_{t}\) and coherently inpaint the correlated depth one under guidance.

\section{Source Frame Selection}\label{sourceFrameSelection}
In a multi-view inpainting system, we first need to sort out a set of source frames from the input RGB-D sequence that can fill in the blanks in \(F_{t}\). This is a quite challenging mission considering the giant quantity and variable qualities of the input frames. In this work, we comprehensively appraise the suitability of each frame to be used as source. 

\subsection{Strategy}\label{localSimilarity}
Given a potential source frame \(F_{i}\), the inpainting accuracy depends on both the image quality itself and the correlations between it and the target frame \(F_{t}\). Therefore, we respectively grade the similarity, the inter-frame distance and the image quality of each \(F_{i}\) to evaluate whether it is suitable or not to inpaint \(F_{t}\). It is reasonable to select these three factors because similarity ensures texture consistency; Larger distance would lead to lower warping accuracy by reducing the amount of matched feature points; as well as blurred image would not only weaken the quality of the inpainted image, but may also trigger mismatches among feature points, which would further lead to high-bias warping results.  

\begin{figure}[t]
  \centering
  \subfloat[]{\includegraphics[width = 0.55\linewidth,height=3.8cm]{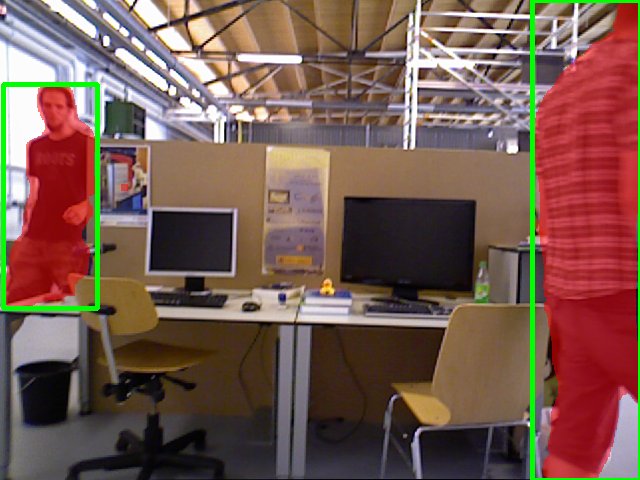}}\
  \subfloat[]{\includegraphics[width = 0.13\linewidth,height=3.8cm]{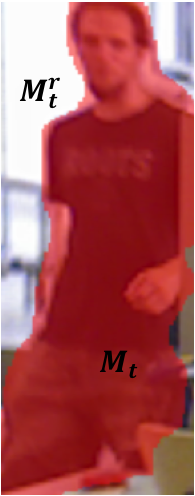}}\
  \subfloat[]{\includegraphics[width = 0.13\linewidth,height=3.8cm]{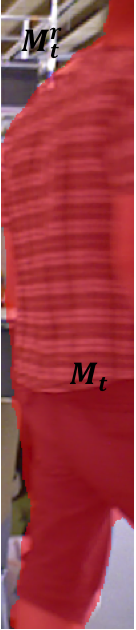}}\
  \caption{(a) The color image of a target frame. Red color indicates completion region and each green box is a minimum circumscribed rectangle of one mask. (b) and (c) Every box is one sub-image and in each one of them, the red region indicates \(M_{t}\) and the left part is \(M_{t}^{r}\)}.
  \label{illu4Patch}
\end{figure}

A certain target frame \(F_{t}\) can contain several masks. Instead of inpainting them simultaneously, we emphasize local similarity and treat each mask \(M_{t}\) on \(F_{t}\) separately. Specifically, we split \(F_{t}\) into a set of sub-images by extracting the minimum circumscribed rectangle for every \(M_{t}\), as shown in Fig.~\ref{illu4Patch}. Each \(M_{t}\) independently gets its own source frames and has not influence on the selections of the others.

We evaluate the suitability \(S(F_{i},M_{t})\) for inpainting \(M_{t}\) with potential source frame \(F_{i}\) by Eq.~\ref{scoreEquation}
\begin{equation}
S(F_{i},M_{t})=\frac{(w_{1}s(F_{i},M_{t})-w_{2}d(F_{i},M_{t}))q(F_{i})}{q(F_{t})},
\label{scoreEquation}
\end{equation}
where \(s(F_{i},M_{t})\) stands for the similarity between \(F_{i}\) and \(M_{t}\); \(d(F_{i},M_{t})\) represents the distance between \(F_{i}\) and \(M_{t}\); as well as \(q(F_{i})\) and \(q(F_{t})\) shows the image quality of \(F_{i}\) and  \(F_{t}\) respectively. \(w_{1}\) and \(w_{2}\) are positive numbers for weight parameters. We use linear normalization to normalize all the factors to \([0,1]\) range. 

In practicing, it is pretty burdensome to calculate the distance term for all the \(F_{i}\) in \(F\) and hence we employ a two-step selection strategy. Namely, we first select the most similar \(n\) source frames by only using the similarity term \(s(F_{i},M_{t}\) and then the \(S(F_{i},M_{t})\) is calculated within these \(n\) frames. 

\subsection{Factors}
We consider the similarity evaluation as a typical image retrieval problem and in this work we use the bag of visual word (BoVW) model to solve it. In practice we use the SIFT feature~\cite{ng2003sift} and vocabulary tree~\cite{GalvezTRO12} for description and searching. The feature points are extracted from unmasked area of the sub-image \(M_{t}^{r}\) as shown in Fig.~\ref{illu4Patch} (b) and Fig.~\ref{illu4Patch} (c).  

For the distance term, we take advantages from the extrinsic parameters of the camera. Since the distance relation between \(F_{i}\) and \(M_{t}\) is the same as that between \(F_{i}\) and \(F_{t}\), we define the distance \(d(F_{s}, M_{t})\) as follows
\begin{equation}
	d(F_{i}, M_{t})=\begin{Vmatrix}r(F_{i}, F_{t})\end{Vmatrix}
+\begin{Vmatrix}t(F_{i}, F_{t})\end{Vmatrix},
\label{disBetween2frames}
\end{equation}
where \(r(F_{i}, F_{t})\) and \(t(F_{i}, F_{t})\) represent the rotation and translation distances respectively between the potential source frame \(F_{i}\) and the target one \(F_{t}\). We set \(d(F_{i}, M_{t})=+\infty\) for unsolvable conditions, which are discarded when doing normalization.

We use gradient based methods~\cite{mittal2012no,bahrami2014fast,ying2015iqa} for image quality assessment, which base their evaluation criteria on the proportion of relatively large gradients in all of them. A higher proportion indicates a larger amount of explicit edges and hence stands for higher image quality. This kind of methods is suitable for our case since we consider edges consistency as one of the factors in MRF (specifically described in section \ref{MRF}) and coherently use gradients in the energy function of it.


\section{Proposals Registration and Combination}\label{registerAndCombine}
So far the source frames have been retrieved, the first thing to do is warping them into the same viewpoint with the target frame. For RGB-D sequence, the classical method to achieve pixel-wise correspondence is to use depth information for calculating the SE(3) transformation. However, a pixel with unknown depth value cannot be transformed and the information loss triggered by such invalid transformation would significantly depress the inpainting accuracy. Hence instead of it we propose to use local homography based warping method in this work.

Given the registered source frames, we need to decide from which one of them the mask pixel should get its value. A naive method is to project all the source frames into the target one and then blending them. However it is easy to trigger blurriness. In this work we propose to use MRF to make optimal choices among sources by considering it as a multi-label problem, similar to~\cite{whyte2009get,thonat2016multi,agarwala2004interactive}. 

We also carry out post refinements for more natural and preciser results after solving the MRF. The entire framework of the proposed warping and combination approach is summarized in algorithm~\ref{WarpAndCombineAlgorithm}. 

\begin{algorithm}[t]
 \caption{Warp and combine multiple proposals}
 \begin{algorithmic}  
 \FOR {Each Mask $M_{t}$ in the target frame}
 	\FOR {Each selected source frame $F_{i}$}
    	\STATE Warp $F_{i}$ into the same viewpoint with the target frame $F_{t}$;
    \ENDFOR
    \FOR {All the color images $I$ of $F_{i}$}
    	\STATE Calculate the average $SSD$ between each $I_{i}$ and $I_{t}$;
        \STATE Create the median image by weighted overlaying each $I_{i}$;
    \ENDFOR
     	\STATE Minimize the energy function of MRF by graph-cut;
    \FOR {Each source frame $F_{i}$}
    	\STATE Compute the transformation matrix $T_{ir}$ between $F_{i}$ and $F_{t}$;
    	\FOR {Each source frame $F_{j} (j\neq i)$}
        	\STATE Compute the transformation matrix $T_{ij}$ between $F_{i}$ and $F_{j}$;
    	\ENDFOR
    \ENDFOR
    \STATE  Pose graph initialization;
    \FOR {Each vertexes $V_{i}$ in the graph}
    	\STATE Assign $T_{ir}$ to $V_{i}$;
        \FOR {Each vertexes $V_{j} (j\neq i)$ in the graph}
        	\IF {$T_{ij}$}
            	\STATE	Assign $T_{ij}$ to the edge $E_{ij}$ that connects $V_{i}$ and $V_{j}$;
            \ELSE 
            	\STATE Disconnect the two vertexes;
            \ENDIF
        \ENDFOR 
    \ENDFOR	
    \STATE Do graph optimization;	
 \ENDFOR
  \STATE Do Poisson image editing;	
 \end{algorithmic}
 \label{WarpAndCombineAlgorithm}
 \end{algorithm}
 
\subsection{Image warping}
A single global homography is the simplest solution to describe pixel-wise correspondences between images. However it is restricted to subject to planar scene or pure rotation motion assumptions. Multi-homography methods~\cite{whyte2009get,liu2016robust} have been employed in various kinds of research to deal with scenes that contain multiple planes, in which the images are divided into several planes and for each of them, an independent homography matrix is calculated. In this work we use the grid based local homography method for warping images taken with freely moving cameras. Whether a local homography can be calculated or not depends on the abundance of matched feature points, hence we use the affine SIFT~\cite{morel2009asift} for points extraction.

\subsection{Multiple Proposals Combination} \label{MRF}
Each source image (for convenience, without ambiguity we hereafter use source image to refer to the color image in the source frame) is considered as a label \(l\) . Our goal is to assign a label \(l_{p}\) for each pixel \(p\) in the mask area. We use the data cost term to represent the cost of assigning \(l_{p}\) to \(p\) and the smooth cost term to encourage avoiding explicit boundaries among distinct proposals. The energy function we would like to minimize it formulated in the form of
\begin{equation}
	\begin{aligned}
		E(l)=&\sum_{p\in\varsigma}(\lambda _{1}T_{1}(p,l_{p})+\lambda _{2}T_{2}(p,l_{p}))\\+&\sum_{(p,q)\in\zeta}\lambda _{3}W(p,q,l_{p},l_{q}),
	\end{aligned}
\label{mrfEnergyFunc}
\end{equation}
where the sum of \(T_{1}(p,l_{p})\) and \(T_{2}(p,l_{p})\) indicate the data cost and \(\varsigma\) represents all the pixels in the sub-image. \(W(p,q,l_{p},l_{q})\) stands for the smooth cost; \((p,q)\) is a pair of neighboring pixels of which we use the 4-neighbor system and \(\zeta\) is the set of all such pairs in the sub-image region. \(\lambda _{1}\), \(\lambda _{2}\) and \(\lambda _{3}\) are weight parameters. It is important to notice that the unmasked pixels are also taken into account as contributions of the total energy since it can serve as constraint for the masked ones via the smooth cost term.   

\(T_{1}(p,l_{p})\) is set to infinity if \(l_{p}(p)\) belongs to mask, representing waiting to be inpainted by in the hole filling section later on. For other cases, we choose \(T_{1}(p,l_{p})\) to have the form
\begin{equation}
	T_{1}(p,l_{p})=\left\{\begin{matrix}
	\left \| I_{l_{p}}(p) -I_{t}(p)\right \|\quad   \ p\in M_{t}^{r}\\ \\
	\left \| I_{l_{p}}(p) -I_{m}(p)\right \|\quad  \ p\in M_{t}
	\end{matrix}\right.,
\label{energyFuncT1}
\end{equation}
where \(I_{l_{p}}(p)\) indicates the value of pixel \(p\) on the proposal \(l_{p}\); \(I_{t}(p)\) is that on the target image and \(I_{m}(p)\) represents the one on the median image, which is considered as a weighted combination of all source images with the rule
\begin{equation}
	I_{m}(p) = \sum_{i\in N}w_{i}I_{l_{i}}(p),
\label{meanImgFunc}
\end{equation}
where \(N\) is the set of source images; \(I_{l_{i}}(p)\) is the value of pixel \(p\) on the proposal \(l_{i}\) and \(w_{i}\) is weight. The weight is imported basing on the fact that the warping accuracies of all the proposals are not equal. Therefore we use the \(w_{i}\) for describing the relative accuracy and define it as 
\begin{equation}
w_{i} =1- \frac{ SSD(I_{l_{i}}(p),I_{t}(p))/n_{i} }{\sum_{j\in N} SSD(I_{l_{j}}(p),I_{t}(p))/n_{j} },
\label{meanImgWeightFunc}
\end{equation}
where \(SSD\) is the sum of square difference between two pixel values; and \(n_{i}\) and \(n_{j}\) are the amounts of overlapped unmasked pixels between the source and target, which is used for normalization. Also we linearly normalize the RGB value to avoid influence caused by illumination changes.

As described earlier, the warping accuracy drops notably when distance is increased between viewpoints. Therefore we choose \(T_{2}(p,l_{p})\) to have the form
\begin{equation}
	T_{2}(p,l_{p})=\exp(\left \| t(I_{l_{i}},I_{t}) \right \|)-1,
\label{energyFuncT2}
\end{equation}
where \(t(I_{l_{i}},I_{t})\) is the translation distance between \(I_{l_{i}}\) and \(I_{t}\); different from Eq.~\ref{disBetween2frames}, we ignore rotation distance since, as mentioned above, in principle, pure rotation would not introduce errors in homography computation. Also this function can provide approximately linear and distinguishable error term. 

For the smooth cost term, we use the same gradient cost function proposed in~\cite{whyte2009get}, where \(W(p,q,l_{p},l_{q})=0\) if \(l_{p}=l_{q}\) and otherwise
\begin{equation}
\begin{aligned}
	W(p,q,l_{p},l_{q})=&\left \| \nabla I_{l_{p}}(p)-\nabla I_{l_{q}}(p) \right \|\\+&\left \| \nabla I_{l_{p}}(q)-\nabla I_{l_{q}}(q) \right \|,
\end{aligned}
\label{smoothCostFunc}
\end{equation}
where \(\nabla I_{l_{p}}(p)\) indicates the gradient of \(I_{l_{p}}\) at pixel \(p\) and so on. We use the Sobel operator to get it.

Finally, to minimize the energy function described in Eq.~\ref{mrfEnergyFunc}, we use the graph-cut algorithm proposed in~\cite{boykov2001fast,kolmogorov2004energy,boykov2004experimental}.

\subsection{Color Adjustment}
The brightness of the same scene may vary among distinct source images because of the different illumination conditions, which will cause obvious contrasts among the boundaries of sources. Therefore, in practice the mask is first expanded to a few more pixels with a dilation filter before solving the MRF and then the Poisson image editing~\cite{perez2003poisson} technique is used for blending the varied lighting.

\subsection{Depth Transformation}
Within previous procedures we have established pixel-wise correspondence between the target frame and the source ones. Now the task is to transform the depth from sources to the target. In principle we achieve it via following the classical "inverse projection, SE(3) transformation and projection" steps. For those source pixels whose depth values are unknown, we set them to blank to represent waiting to be inpainted in the hole filling section later on. Namely, for each source frame a transformation matrix \(T\in \mathbb{R}^{4\times 4}\) should be calculated.

However, this is only part of the story. Another problem is that the estimation of pairwise transformation is not such accurate. Also considering that information for inpainting is collected from multiple proposals, it is easy to trigger inconsistencies on the inpainted depth image. Therefore, we implement a global optimization process to get the preciser transformation matrices by modeling it into a graph optimization problem. Specifically, we take the pose of the target frame as the origin and its camera coordinate as the world one. Then the vertexes in the graph can be set to the transformation matrices \(T_{sr}\) from each source frame to the target. Coherently, the edges connecting pairwise vertexes are conditionally either set to the transformation matrices \(T_{ij}\) between the two source frames if the \(T_{ij}\) can be computed or left as blank to represent disconnections. We employ the g2o framework~\cite{kummerle2011g} for implementation.


\section{Hole filling}\label{holeFilling}
Haven finished the multiple proposals combination, we can now proceed to the final step to cope with the still existing holes on the target frame. For the color image, a multi-view exemplar based inpainting method is proposed. Also, the inpainted color image will serve as guidance for completing the corresponding depth one.

\subsection{Color Image Inpainting}
Exemplar based inpainting methods basically synthesize values for the mask from the source by minimizing an energy function describing the similarity between them. In this work, we base our multi-view exemplar inpainting approach on the method proposed in~\cite{kawai2009image,kawai2012image} and define the energy function in the form of
\begin{equation}
\begin{aligned}
E=\sum_{p_{i}\in \phi, p_{j}\in \Phi }SSD(p_{i},p_{j}),
\end{aligned}
\label{exemplarInpaint}
\end{equation}
where, the same as the general definitions in image inpainting work, \(\phi\) is the boundary area of the mask; \(\Phi\) is the source area from where the information is gotten; \( p_{i}\) and \(p_{j}\) respectively indicate pixels in \(\phi\) and \(\Phi\); and \(SSD\) represents the patch similarity in the form of
\begin{equation}
\begin{aligned}
SSD(p_{i},p_{j})=&\sum_{s\in \omega} \left \|I(p_{i+s})-\alpha _{p_{i}p_{j}}I(p_{j+s})\right \|\\+&\left \| \nabla I(p_{i+s})-\nabla I(p_{j+s}) \right \|,
\end{aligned}
\label{exeSSD}
\end{equation}
where \(\omega\) is the patch size; \(s\) is a shift vector used for traverse all the pixels within \(\omega\); \(I(p_{i+s})\) and \(I(p_{j+s})\) indicate values of pixel \(p_{i+s}\) and \(p_{j+s}\); \(\nabla I(p_{i+s})\) and \(\nabla I(p_{j+s})\) are gradients used as constraints for preserving texture consistency. \(\alpha _{p_{i}p_{j}}\) is a parameter used for dealing with brightness changes~\cite{kawai2009image}, which is defined as
\begin{equation}
\alpha _{p_{i}p_{j}}=\sqrt{\frac{\sum_{s\in \omega} I^{2}(p_{i+s})}{\sum_{s\in \omega} I^{2}(p_{j+s})}},
\label{brightRatio}
\end{equation}

For multi-view inpainting, we expand the definition of \(\Phi\) as the unmasked area in the single image to that in both the target image and the source ones used for proposals combination. It is worth mentioning that the source images used in our approaches are the warped ones as described in section~\ref{MRF} to ensure texture consistency between the inpainted area and the original unmasked one. It is reasonable to do so since the warped images are in the same viewpoint with the target one, which means that we are seeking information from a spatially consistent region rather than several independent counterparts. 

\subsection{Guided Depth Image Inpainting}
Depth images are usually low textured and propagation based methods are hence applicable. Similar to~\cite{miao2012texture}, we extract edges from the color image as guidance and coherently divide the masked pixels of the depth image into either smooth or edge classes. Specifically, a pixels would be classified into edge class if its color correspondence is edge and smooth otherwise. Propagations are processed separately on each class.

\begin{figure}
\centering
  \includegraphics[width=0.5\textwidth]{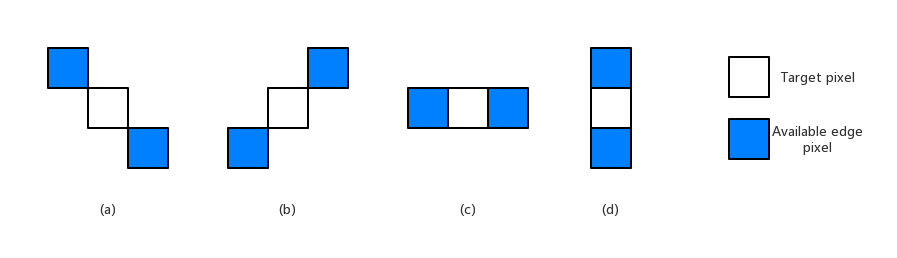}
  \caption{Creditable distribution samples for edge class}
\label{edgeCondition}
\end{figure}

A given pixel can reckon its value from propagation only if its neighbors can provide enough information. In this work we use the 8-neighbor system and define respective rules for smooth and edge classes to evaluate the conditions of their neighbors. For the smooth class, a creditable neighbor should has no less than 4 available pixels that also belong to the smooth; and for the edge one, a neighbor region is reliable if the distribution of available edge pixels within it satisfies one of the conditions shown in Fig.~\ref{edgeCondition}. 

The mask area is iteratively inpainted until convergence. In each iteration, a pixel is either set to the propagated value provided satisfying the rules above or skipped otherwise. An edge pixel would be consider as mis-masked and move to the smooth class if it still cannot be assigned value after several loops. For propagation we use the Laplace equation in its discrete form
\begin{equation}
I(p,t+1)=\frac{\sum_{p'\in \mu _{8}(p)} \kappa (p')I(p',t)}{\sum_{p'\in \mu _{8}(p)}\kappa (p')},
\label{PDE}
\end{equation}
where \(I(p,t+1)\) is the value of pixel \(p\) at time \(t+1\); \(\mu _{8}(p')\) indicates the 8-neighbor pixels of \(p\); and \(\kappa (p')\) is an indicator function which is set to 1 if the pixel is in the same class with \(p\) and 0 otherwise.


\begin{figure*}[t]
\centering
  \includegraphics[width=\textwidth]{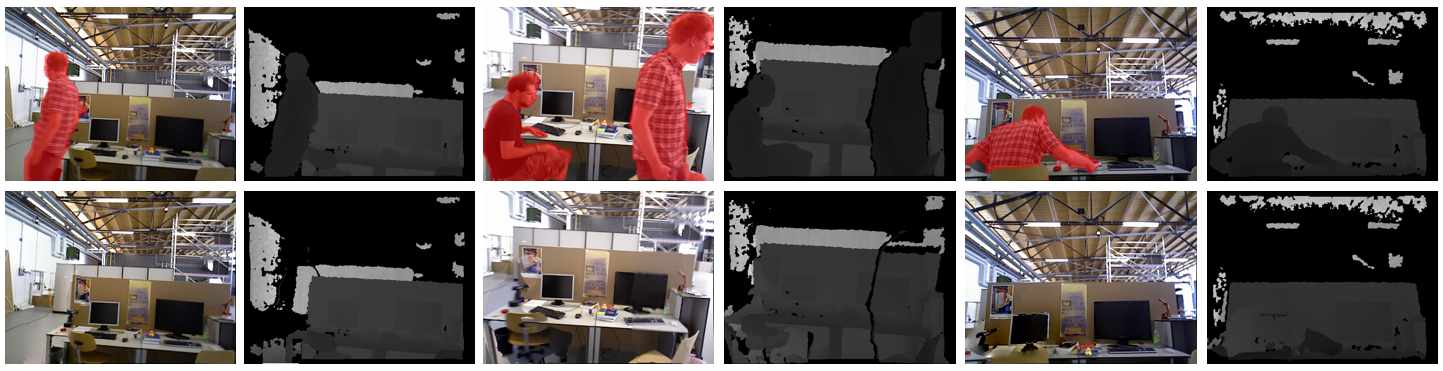}
  \caption{Testing results from different datasets. The first row indicates the original frames and the second one is the inpainted results with our method. The first two columns come from the freiburg3-walking-xyz dataset; middle two columns: freiburg3-walking-rpy; last two columns: freiburg3-walking-halfsphere.}
\label{result}
\end{figure*}

\begin{figure*}[t]
\centering
  \includegraphics[width=\textwidth]{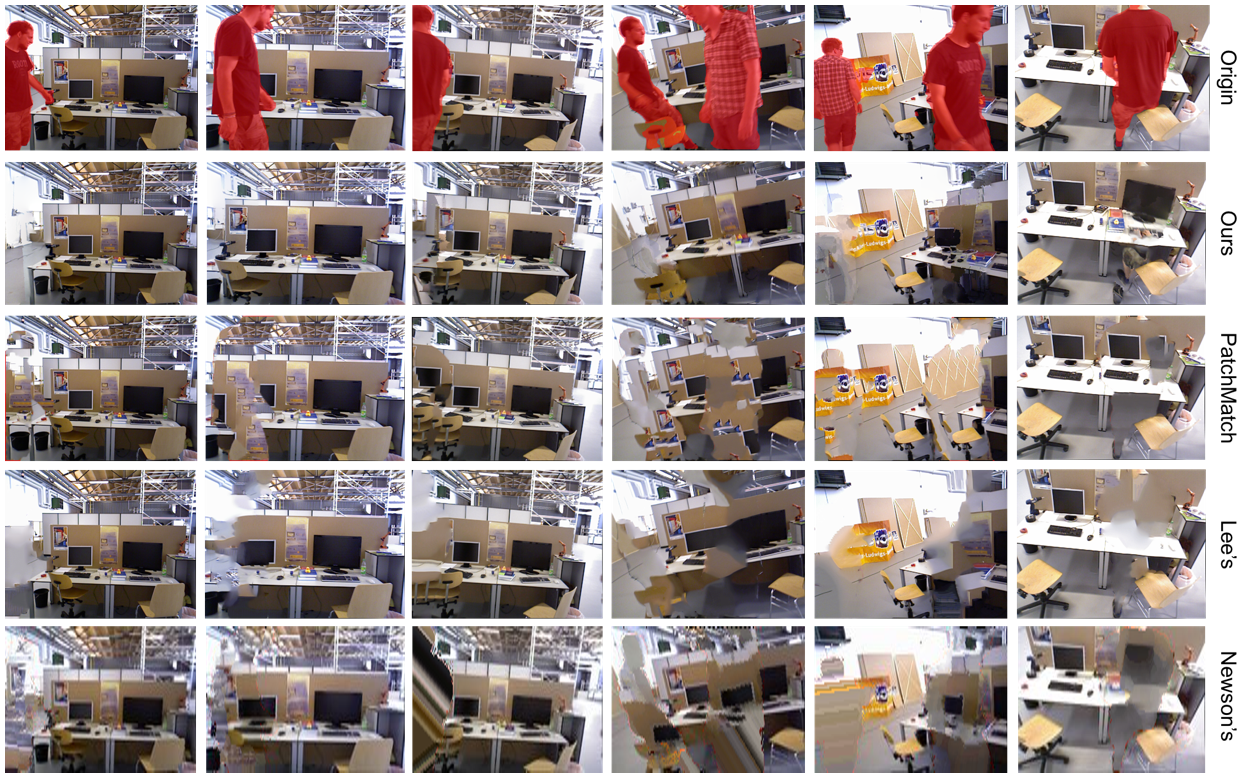}
  \caption{Compare with other methods. Each column shows one image inpainted by different methods. The first and second rows show the origin images and inpainting results with our method. The third and fourth rows present the results of single image inpainting methods respectively proposed by Barnes \textit{et al.}~\cite{barnes2009patchmatch} and Lee \textit{et al.}~\cite{lee2016laplacian}. The last row is the results of the video inapinting algorithm proposed by Newson \textit{et al.}~\cite{newson2014video}.
  }
\label{compare}
\end{figure*}

\section{Results and Comparisons}\label{resultAndCompare}
We test our approach on different datasets from the TUM RGB-D benchmark~\cite{sturm12iros} (freiburg3-walking-rpy, freiburg3-walking-xyz and freiburg3-walking-halfsphere), in which two persons randomly walk across the scenes and the cameras are also on moving. The "human" class is defined as undesired. It is important to notice that these three datasets are totally different because of the distinct camera motions although they present similar scenes. These datasets are pretty challenging for being highly textured. For each dataset we use the first 800 frames and one target frame within it as input. The results are shown in Fig.~\ref{result}. 

 \begin{figure}[t]
\centering
  \includegraphics[width=0.45\textwidth]{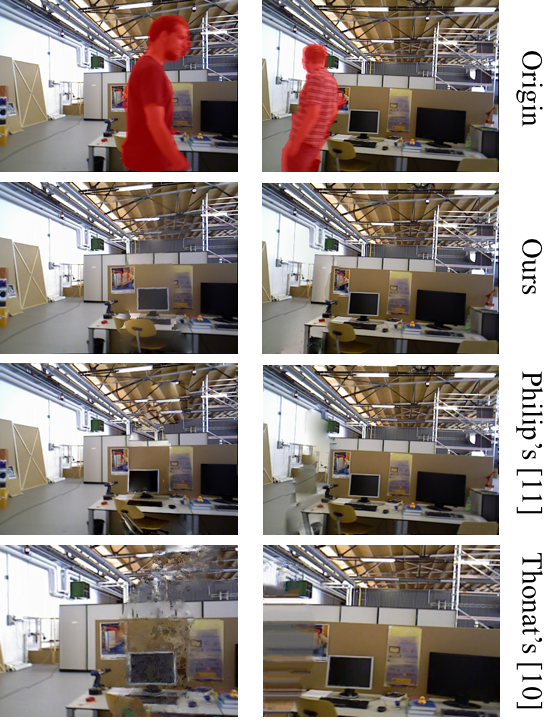}
  \caption{Compare with other multi-view inpainting approaches. Our method can effectively preserve texture consistencies (\textit{e.g.} the upper bound of the white board and the ceiling). }
\label{multiComp}
\end{figure}

\subsection{Application Example}
\begin{figure}[t]
\centering
  \includegraphics[width=0.8\linewidth]{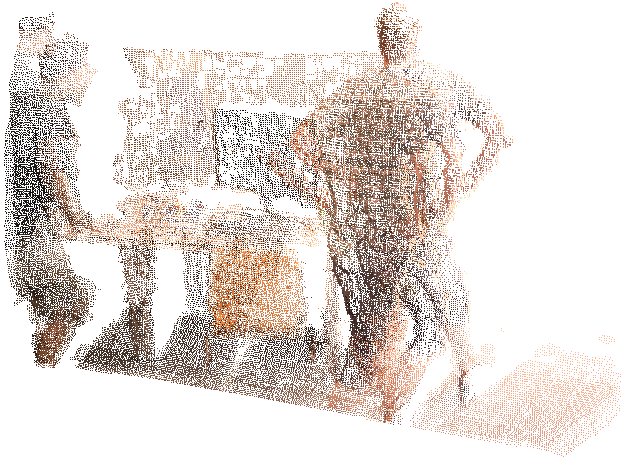}
  \includegraphics[width=0.8\linewidth]{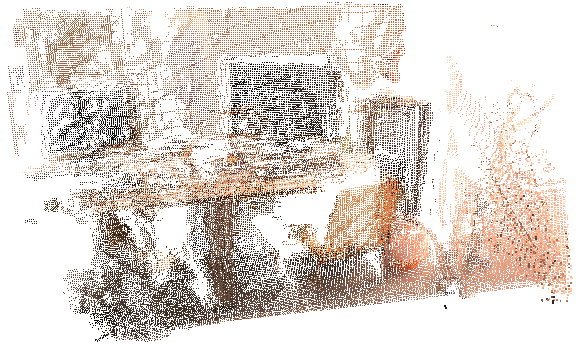}
  \caption{Application example of our proposed algorithm. Above: point clouds projected from original images. Below: the counterpart after inpainting with our method.}
\label{usageExample}
\end{figure}

Hereby we present an application example of our work. For a set of selected frames, the fine-tuned PSPNet~\cite{zhao2017pyramid} is used to detect and draw masks on the "human" class. Also we expand the masks to a bit more pixels by dilation in order to cope with unmasked edges~\cite{li2018UR}. We use MeshLab~\cite{cignoni2008meshlab} to present the final results as shown in Fig.~\ref{usageExample}. 

\subsection{Comparisons}
As far as we know, similar work that semi-autonomously inpainting RGB-D sequences barely exists and therefore we could only compare our algorithm with other color image inpainting techniques. Specifically, we compare our method with other three approaches on all the three datasets, two of which are single image inpainting methods~\cite{barnes2009patchmatch,lee2016laplacian} and the other one is proposed for video inpainting~\cite{newson2014video}. The results are presented in Fig.~\ref{compare}. Evidently the single image inpainting approaches cannot perfectly handle large blanks. 
As for the video inpainting method, we have to down-sample the images and use the neighboring 100 frames of the target as input because of its extremely high time cost. We also present comparisons among our methods and those proposed in~\cite{thonat2016multi,philip2018plane}, as shown in Fig.~\ref{multiComp}.


\section{Conclusions and Discussion}\label{conclusionAndDiscussion}
We have introduced a multi-view based method for inpainting RGB-D sequences. Experiment shows the improvements of our method over the existing ones. However, like other homography based approaches, our method is easy to suffer from the sparseness of feature points when handling large baseline conditions. Also the segmentation accuracy would significantly infect the inpainting quality since the undesired objects might be re-introduced into the target frame provided poor segmentation. 

For future work, other methods like those employing grid optimization can be explored to deal with large baseline conditions~\cite{lin2015adaptive}. Also more suitable source selection strategy could be designed to avoid the current demand on weights adjustment. Another line of interest is, besides segment out the movable objects themselves (like the human), other objects which are passively moved (like the chair) can also be taken into account. 


\section*{Acknowledgement}
We thank Norihiko Kawai for providing the source code of~\cite{kawai2012image}; and the authors of~\cite{newson2014video,thonat2016multi,philip2018plane} for their kind help on making comparisons.

{\small
\bibliographystyle{ieeetr}
\bibliography{reference}
}
\end{document}